\documentclass[lettersize,journal]{IEEEtran}

\usepackage{amsmath,amsfonts}
\usepackage{algorithmic}
\usepackage{algorithm}
\usepackage{array}
\usepackage[caption=false,font=normalsize,labelfont=sf,textfont=sf]{subfig}
\usepackage{textcomp}
\usepackage{stfloats}
\usepackage{url}
\usepackage{verbatim}
\usepackage{graphicx}
\usepackage{cite}
\usepackage{multirow}
\usepackage{threeparttable}

\usepackage{hyperref}
\hypersetup{
colorlinks=true,
linkcolor=blue,
citecolor=blue,
urlcolor=black
} 
\usepackage{booktabs}

\DeclareMathOperator*{\argmin}{arg\,min}

\begin{document}

\title{Beyond Gait: Learning Knee Angle for Seamless Prosthesis Control in Multiple Scenarios}

\author{Pengwei Wang, Yilong Chen, Wan Su, Jie Wang, Teng Ma, Haoyong Yu ~\IEEEmembership{}

\thanks{Pengwei Wang, Yilong Chen, Wan Su, Jie Wang are with the Department of Mechanical Engineering, National University of Singapore,  Singapore 117575, Singapore.}
\thanks{ Teng Ma and Haoyong Yu are with the Department of Biomedical Engineering, National University of Singapore, Singapore 119077,
 Singapore.}
}

\maketitle

\begin{abstract}
Deep learning models have become a powerful tool in knee angle estimation for lower limb prostheses, owing to their adaptability across various gait phases and locomotion modes. Current methods utilize Multi-Layer Perceptrons (MLP), Long-Short Term Memory Networks (LSTM), and Convolutional Neural Networks (CNN), predominantly analyzing motion information from the thigh. Contrary to these approaches, our study introduces a holistic perspective by integrating whole-body movements as inputs. We propose a transformer-based probabilistic framework, termed the Angle Estimation Probabilistic Model (AEPM), that offers precise angle estimations across extensive scenarios beyond walking. AEPM achieves an overall RMSE of 6.70 degrees, with an RMSE of 3.45 degrees in walking scenarios. Compared to the state of the art, AEPM has improved the prediction accuracy for walking by 11.31\%. Our method can achieve seamless adaptation between different locomotion modes. Also, this model can be utilized to analyze the synergy between the knee and other joints. We reveal that the whole body movement has valuable information for knee movement, which can provide insights into designing sensors for prostheses. The code is available at \href{https://github.com/penway/Beyond-Gait-AEPM}{https://github.com/penway/Beyond-Gait-AEPM}.
\end{abstract}


\section{Introduction}
\IEEEPARstart{L}{ower} limb amputation profoundly affects both the physical capabilities and emotional health of individuals, diminishing their quality of life \cite{amtmann2015health}. In 2005, 1.6 million Americans experienced limb loss, a number that is expected to double by 2050 \cite{ziegler2008estimating}. This issue, significant both in the United States and concurrently across the globe, emphasizes its magnitude as a worldwide concern. Consequently, prostheses and assistive devices are crucial to improving life quality and social participation for people with limb disabilities. Active lower limb prostheses, equipped with a power source, provide enhanced support for amputees in tasks such as climbing stairs or slopes, and getting up from a sitting position, compared to passive or semi-passive prostheses \cite{windrich2016active}. The enhanced mobility offered by these active devices hinges on sophisticated control systems that dynamically adapt to the user's movements. The success of these systems depends on accurately predicting the knee angles, which enables the nuanced adjustment of the prosthesis in real time to match the user's various activities.

Current methods for knee prosthesis control mainly focus on gait cycles. Most approaches leverage Finite State Machine (FSM), which segments a gait cycle into multiple sections \cite{sup2008design} and utilizes impedance control strategy \cite{hogan1984impedance}. The estimated knee motion relies on impedance functions. The transition between sections, or states, is governed by the detection of specific gait events, such as heel strikes or toe-offs. This process guarantees that the impedance control aligns with the natural progression of the gait cycle. The control parameters vary between different gait sections, which are categorized into various locomotion modes, such as level walking, ramp traversal, and stair climbing. As a result, the number of parameters requiring fine-tuning for each user, as well as for each new mode of locomotion, significantly increases. Future efforts in impedance control are primarily directed towards the development of automatic tuning methods \cite{huang2016cyber, simon2014configuring}. While these improvements diminish the requirement for tuning by experts, FSM remains constrained by its predetermined modes, resulting in a notably deficient generalization capability. Furthermore, the fundamental nature of impedance control precludes its ability to facilitate users in activities involving elevation, such as ascending stairs or transitioning between sitting and standing positions.

\IEEEpubidadjcol

\begin{figure*}[h]
    \centering
    \includegraphics[width=0.95\linewidth]{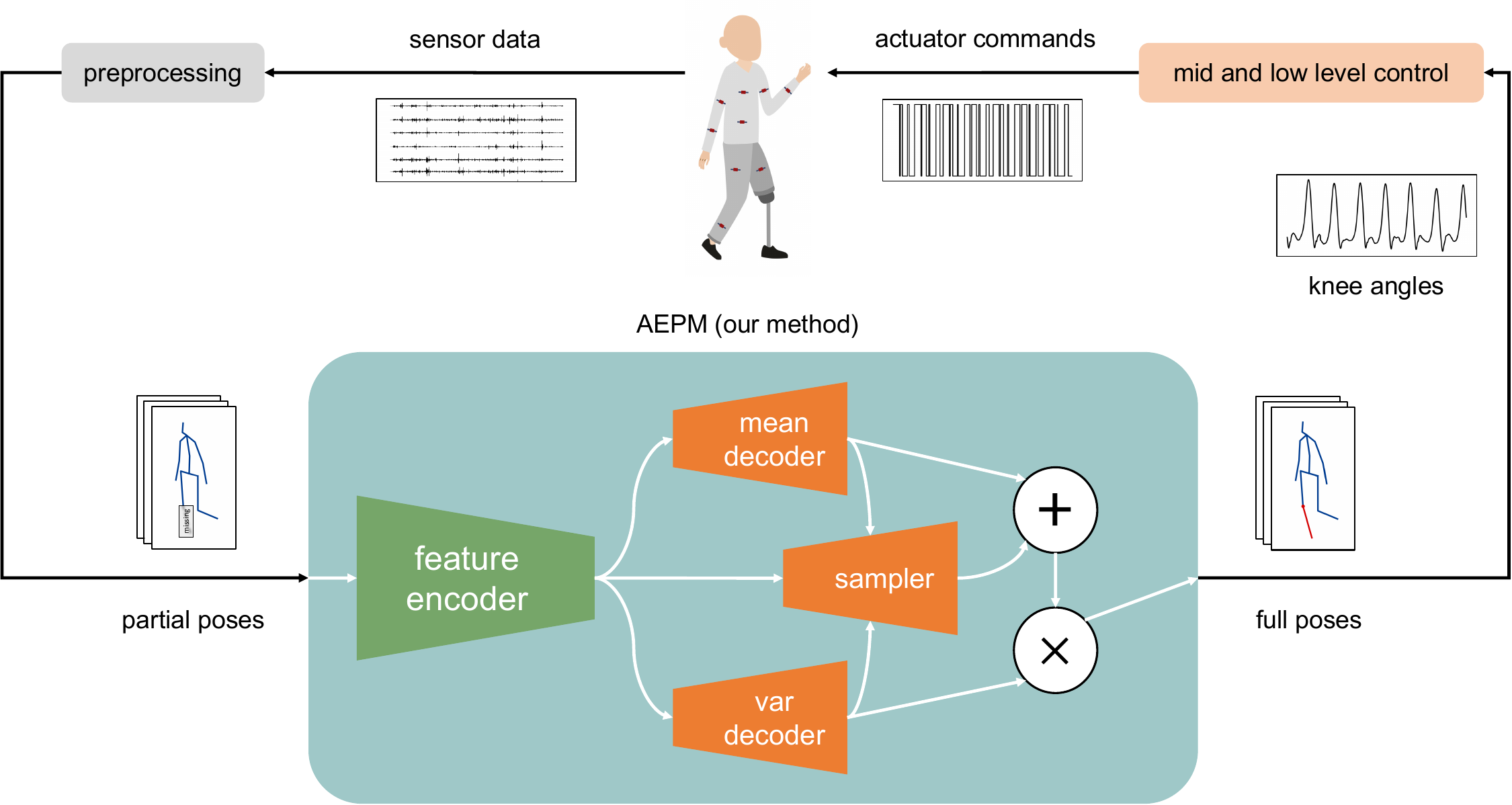}
    \caption{Overview of the proposed method, and potential pipeline when integrated in the control system. Our method reconstructs the partial poses into full poses}
    \label{fig:overview}
\end{figure*}

Compared to FSM, methods of continuous angle estimation can better ensure the smoothness of prostheses' movements. Traditional methods involve a functional mapping between sensors (force sensor\cite{dong2023low}, inertial measurement unit (IMU) \cite{quintero2017real}, EMG \cite{shen2023simultaneous}) and gait phases to estimate knee angle. Moreover, with the development of deep learning, such methods have been introduced into this control issue, enhancing the precision of gait phase detection. Network architecture types including artificial neural network (ANN) \cite{stetter2020machine}, convolutional neural networks (CNN) \cite{kang2021real}, graph convolutional networks (GCN) \cite{wu2021gait}, long short-term memory (LSTM) \cite{cai2023gait,guo2023transferable,jeon2024bi} and attention based model \cite{chen2024novel} have been previously investigated to estimate gait phase. Some approaches bypass gait phase estimation and focus directly on generating knee angles or torques instead. For instance, Nuesslein et al. use the temporal convolutional network (TCN) to generate stance states and joint torques end to end \cite{nuesslein2024deep}. A transferable multi-modal fusion (TMMF) is proposed by Guo et al. to perform a continuous prediction of joint angles by using the maximum mean discrepancy (MMD) and LSTM network \cite{guo2023transferable}. Ding et al. proposed an attention-based method for estimating knee angles, using sensors on healthy contralateral limbs\cite{ding2023deep}. However, whether using the method of estimating the gait phase or directly estimating joint angles or torques, their applicability is limited to gait-related scenarios, which are essentially rhythmic movements. In reality, a significant portion of daily activities encompasses various non-rhythmic motions (e.g. standing up, sitting down, etc.).

In this study, we develop a novel approach to predict knee joint angles, utilizing the simultaneous state of other joint angles throughout the body. Human motion exhibits interdependence between the lower limbs and other body parts to achieve stability and energy efficiency \cite{seyfarth2022whole}. For example, neurological research supports the consensus that the central pattern generator (CPG) within the spinal cord's neuronal circuitry coordinates various body parts to work during locomotion \cite{sharbafi2017locomotion}. This inherent coordination offers valuable insights, suggesting that knee motion can be inferred from the dynamics of other body regions. 
To better capture the interdependence among joints, we utilize a transformer-based model. Transformers, initially developed for natural language processing tasks \cite{vaswani2017attention}, have achieved notable success across a diverse set of fields. Recently, within the field of computer graphics, transformers have demonstrated their utility in a variety of tasks related to human pose, such as 3D motion prediction \cite{aksan2021spatio, martinez2021pose} or 3D pose estimation in video \cite{zheng20213d, zhang2022mixste}. To our best knowledge, our method represents the first application of transformer model for angle estimation in prostheses.

Our method -- Angle Estimation Probabilistic Model (AEPM) utilizes a modified Mixed Spatial-Temporal Encoder (MixSTE) \cite{zhang2022mixste}, an implementation of transformer to relate the time and spatial behavior of multiple remaining joints and encode the information. Then, we use a multi-modal setting to estimate the probability distribution of knee angle conditioned on the motion of remaining joints. While the test set includes non-traditional gait motions such as jumping and walking sideways, our method can still achieve a good estimation. The contribution of this work is summarized as follows:

1. A deep learning method is proposed for knee position estimation with seamless motion transition without explicit intention classification and irrespective of the action mode.

2. Experimental results illustrate that the proposed method achieves high prediction accuracy for both rhythmic and non-rhythmic movements across a wide range of daily life scenarios.

3. Leveraging the model, we demonstrate that the whole-body movement has rich information for the knee movement.

\begin{figure*}[h]
    \centering
    \includegraphics[width=0.95\linewidth]{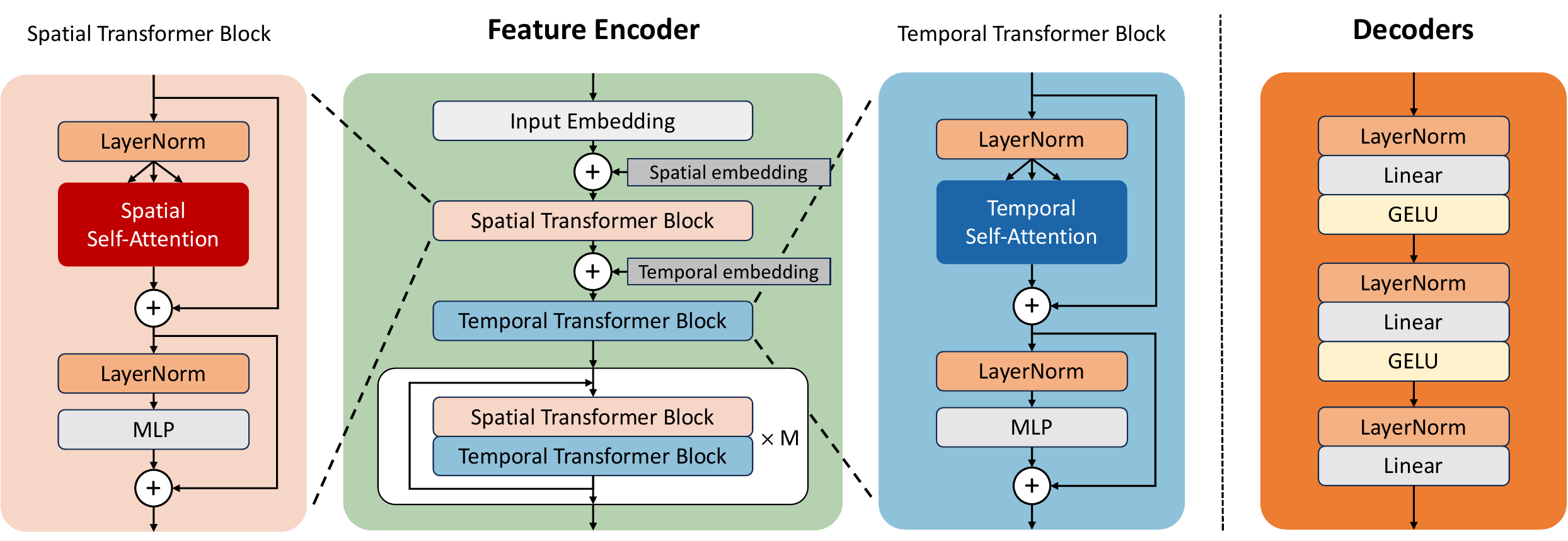}
    \caption{
    Detailed structure of Angle Estimation Probabilistic Model (AEPM). The encoder structure mainly consists of a stack of spatial and temporal transformer blocks, basically following the setting of \cite{vaswani2017attention, zhang2022mixste}. The spatial and temporal self-attention apply the self-attention on spatial and temporal dimensions respectively. The decoders are all Multi-Layer Perceptron which share a similar structure, consisting of three linear layers.
    }
    \label{fig:structure}
\end{figure*}

\section{Methods}

\subsection{Overview}
The core issue explored in this paper is the generation of missing knee angles to control the prosthesis based on the simultaneous state of other joint angles throughout the body. The pipeline of our method, as depicted in Figure \ref{fig:overview}, outlines the proposed sequence of operations from the initial acquisition of sensor data to the final actuation of prosthetic lower limbs.

Notably, this article does not cover capturing current human postures through sensors. Nevertheless, it is fortunate that pertinent studies have successfully navigated this challenge. For instance, real-time postures for each joint can be obtained by implementing a preprocessing routine proposed by Yinghao Huang et al.\cite{huang2018deep} on IMU data derived from diverse body joints.

\subsection{Problem Definition}

The ultimate aim is to estimate the knee angle \(k\) from the partial pose \(\bar{x}\). To do this, our model, Angle Estimation Probabilistic Model (AEPM), recovers the full pose \(\hat{x}\) from partial pose \(\bar{x}\), then the knee angle can be easily retrieved by selecting the corresponding channel. The model estimates a distribution of \(\hat{x}\) similar to the method in \cite{mao2023leapfrog}. First, the partial pose \(\bar{x}\) is encoded by feature encoder \(f_{enc}\) into a latent representation \(c\):
\begin{align}
    c = f_{enc}(\bar{x}).
\end{align}

Then, the mean and variance are decoded from the representation:
\begin{align}
    \mu &= f_\mu(c), \\
    \sigma &= f_\sigma(c).
\end{align}

A sampler is subsequently employed to draw from this distribution, generating a set of samples \(\mathbb{S}\) which are further re-parameterized into final predictions:
\begin{align}
    \{S_i\}_{i=0}^{N} &= \mathbb{S} = f_S(c,\mu,\sigma), \\
    \label{eq:s}
    \hat{x}_i &= \mu + \frac{\sigma}{std(\mathbb{S})} \times S_i.
\end{align}

\subsection{AEPM}
Overall, AEPM is an encoder-decoder-based model. The input \(\bar{x}\) initially enters the feature encoder, where it undergoes feature extraction in both spatial and temporal dimensions and maps these features to the latent space. Based on the probabilistic model mentioned above, features extracted from the latent space undergo decoding through three specialized decoders, each dedicated to generating specific outputs: mean, variance, and N samplers. Finally, following (\ref{eq:s}) mentioned above, the model outputs N sets of predictions.

Next, the encoder and decoder's architecture, as shown in Figure \ref{fig:structure}, is thoroughly explained.

\subsubsection{Feature Encoder}
The feature encoder consists of multiple layers of Spatial and Temporal Transformer Blocks (STB, TTB) \cite{zhang2022mixste}, each one consists of a self-attention block, applying spatial-wise or temporal-wise attention and an MLP block, which is the same setting with \cite{vaswani2017attention}. The initial dimension of input \(x\) is \(b \times l \times n \times 3\), where \(b\) represents the batch size, \(l\) the number of frames, and \(n\) the number of joints. The features are then mapped to a higher dimension of \(b \times l \times n \times d\) through the input embedding layer, which is a single linear layer. After integrating a learnable Spatial Embedding, the features are then fed into the Spatial Transformer with dimensions \( (b \times l) \times n \times d\). Similarly, features are added with learnable Temporal Embedding before being input into the Temporal Transformer for feature extraction. Distinctly, the features will be input in the dimension of \( (b \times n) \times l \times d\). 
Subsequently, features are processed by \(M\) layers of STBs and TTBs.

\subsubsection{Decoder}
Decoders are a set of Multi-Layer Perceptron (MLP) to reduce the dimension of the features from the encoder. The three specialized decoders are very similar in structure. The only difference lies in the output dimensions: the decoders for mean output dimension of \(b \times l \times n \times 3\), and variance have the output dimension of \(b \times l \times n \times 1\), whereas the decoder for the sampler has an output dimension of \(b \times l \times n \times (N \times 3)\), where \(N\) represents the desired number of samplers to be output.

\subsection{Losses}
A two-stage training strategy is used for AEPM to ensure stable training and accurate variance estimation. In the first stage, the Mean Square Error (MSE) is calculated between all samples generated and the ground truth, as is shown in (\ref{eq:ls1}). However, this initial stage often results in the model uniformly predicting minimal variance, inadequately capturing the variability inherent in different poses, which is a critical aspect of our probabilistic framework. This limitation prompted a refined approach in the subsequent stage, where MSE calculation is restricted to the optimal result, as is shown in (\ref{eq:ls2}). This adjustment empirically facilitates a more accurate estimation of variance. 

Additionally, our findings indicate that reconstructing the full pose and calculating the loss based on this comprehensive approach results in superior outcomes compared to focusing solely on the knee angle. This improvement likely stems from the model's enhanced overall understanding of the pose.
\begin{align}
    \label{eq:ls1}
    \mathcal{L}_{S1} &= \frac{1}{N}\sum_i \|x - \hat{x}_i\|_2, \\
    \mathcal{L}_{S2} &= \min({\|x-\hat{x}_i}\|_2).
    \label{eq:ls2}
\end{align}
The discontinuity in the operation of selecting the minimum is addressed by re-parameterizing \(\mathcal{L}_{S2}\) to directly use the sample that minimizes the distance to the ground truth.
\begin{align}
    \mathcal{L}_{S2} = \|x-\hat{x}_j\|_2, j = \argmin_i(\|x-\hat{x}_i\|_2).
\end{align}

\section{Experiments}

\subsection{Datasets}
Two large-scale motion capture datasets are used in the experiments, Human3.6M \cite{ionescu2013human3} and CMU mocap database \cite{cmu_graphics_lab_carnegie-mellon_2003}. In both datasets, the right knee is masked and predicted.

\textbf{Human3.6M}, the more structured one, contains 3D human poses and corresponding images captured at 50Hz, performed by 7 actors with 15 daily scenarios, including discussion, taking photos, talking on the phone, etc. It offers accurate 31 3D joint angles represented in exponential space and a whole-body translation and rotation. In our experiment, we select 14 distinct joints and the whole body rotation, coupled with the masked knee joint to form a \(16 \times 3\) matrix. The joint definition is shown in Figure \ref{fig:attn_skeleton}. Subject 5 (S5) is designated as our test set, with the remaining data forming the training set.

\textbf{CMU mocap database} presents a broader range of movements across 144 subjects, initially captured at 120Hz and uniformly down-sampled by us to 60Hz. It includes 30 joints measured in Euler angles, complete with whole-body translation and rotation. It covers a wide range of movements, from dancing and jumping to unconventional gait styles. In this dataset, we choose 13 out of 30 joints, incorporating the masked knee angle and whole body rotation for analysis, totally 14 joints. The joint definition is similar with Human3.6M, only the 10 and 13 joint angle pointing from neck to shoulders are removed, as they are fixed in this dataset. For the test set, 11 subjects are selected covering multiple movements.

To ensure our model's generalization capability and robustness among different subjects, we opted not to include data from test subjects in our training dataset. This decision reflects the practical difficulty of collecting pre-disability activity data from patients before they require a prosthesis

\subsection{Evaluation Metric}

The Root Mean Squared Error (RMSE), expressed in degrees, serves as the evaluation metric for assessing the accuracy between predicted knee angles and the ground truth. Our evaluation includes calculating both an overall mean RMSE and the best (minimum) RMSE value among all the predicted values. To ensure a fair evaluation, we shift the analysis window by one frame sequentially, calculating the error using only the last frame's predicted value to prevent the model from considering future information.

\subsection{Implementation Details}

In our experiment, a total 16 joints from Human3.6M and 15 joints from the CMU mocap dataset are selected. Before feeding these datasets into the model, we mask the right knee joint information by setting their values to zero.
 
For the model architecture, we set the number of samples for the decoder (\(N\)) as 10 and the number of encoder layers is adjusted to 4 for Human3.6M and 5 for the CMU mocap database. The number of frames (l) is 25 for Human3.6M and 30 for CMU mocap, in other words, the time window is 0.5 seconds for both datasets. The embedded dimension in the encoder (\(d\)) is specified as 32.

In training, we use AdamW \cite{loshchilov2017decoupled} optimizer, with a learning rate of 0.001 and a batch size of 128. We assign the epochs to 500, but the experiment usually early stop when there is no sign of improvement. The shift epoch from the first stage training to the second one is set as 30 and 3 respectively.

\subsection{Performance Analysis}
To address several critical inquiries concerning our method's hypothesis and design, two experiments are designed. The questions under scrutiny are: the necessity of a dual-stage training approach, the imperative of punishing on all joints in calculating loss, the feasibility of attaining comparable performance with a reduced set of joints, and the implications of applying our methodology solely to lower limb joints.

\subsubsection{Ablation Study on Losses}
To prove the effectiveness of our training strategy, three more training losses are used for comparison. Firstly, the two-stage training with loss is applied only on the target knee angle. Additionally, we examined the effects of employing exclusively the first stage loss \(\mathcal{L}_{S1}\) or second stage loss \(\mathcal{L}_{S2}\), compared against the complete two-stage training schema. Detailed outcomes of this examination are elucidated in Section \ref{sec:ablation results}.

\subsubsection{Comparison in Different Joint Quantities}
In extending our analysis, we explored the model's performance across varied joint configurations beyond the initial 15-joint setting utilized for the Human3.6M dataset. This exploration encompassed a minimal 3-joint setting targeting lower-limb joints (joint 1, 6, 7), an augmented setting incorporating an additional hip to lower-back angle (joint 1, 6, 7, 11), and an 8-joint configuration that further integrates whole-body rotation, neck, and shoulder angles (joint 0, 1, 6, 7, 11, 13, 17, 25). These configurations were methodically selected to evaluate performance across a spectrum from lower limb specificity to reduced whole-body dynamics. Results of this experiment are included in Section \ref{sec:joint quanti}.

\begin{table*}
    \centering
    \caption{
    Knee angle estimation results of AEPM on the test set of Human3.6M Dataset \cite{ionescu2013human3} with different scenarios
    }
    \begin{tabular}{lccc *{2}{p{1.3cm}} ccc}
         \toprule
         Movement&  directions&  discussion&  eating&  \centering{greeting} &  \centering{phoning} &  \centering{posing} &  purchases& sitting\\
         \midrule
         mean RMSE (degree) &  3.19& 5.44& 2.71& \centering{4.81} & \centering{10.81} & \centering{8.19} & 4.94& 10.24\\
         best RMSE (degree) &  0.76& 2.08& 0.65& \centering{1.38} & \centering{5.33} & \centering{2.95} & 1.40& 4.11\\
         \midrule
         Movement&  sitting down&  smoking&  taking photo&  \centering{waiting} & \centering{walking} & \centering{walking dog} &  walking together& average\\
         \midrule
         mean RMSE (degree) &  15.15& 5.53& 10.71& \centering{5.53} & \centering{3.45} & 6.08& 3.66& 6.70\\
         best RMSE (degree) &  7.83& 3.14& 4.21& \centering{2.43} & \centering{0.92} & 2.16& 1.00& 2.69\\
         \bottomrule
    \end{tabular}
    \label{tab:res_h36m}
\end{table*}

\begin{table*}
    \centering
    \caption{Knee angle estimation results of AEPM on the test set of CMU mocap Database \cite{cmu_graphics_lab_carnegie-mellon_2003} with different scenarios}
    \begin{tabular}{lcccccc}
         \toprule
         Movement &  Walk and Run  & General Subject & Walk on uneven terrain & Human Interaction &  Basketball & Average \\
         Subject number & (37, 45, 46, 47) & (15, 141, 143) & (3) & (21) & (6) & \\
         \midrule
         mean RMSE (degree) &  7.40 & 6.68 & 14.71 & 8.43 & 6.44 & 8.73\\
         best RMSE (degree) &  1.78 & 3.71 & 7.77 & 3.23 & 2.32 & 3.76\\
         \bottomrule
    \end{tabular}
    \label{tab:res_cmu}
\end{table*}

\section{Results and Discussion}

\subsection{Test Results}

The test results for Human 3.6M are shown in Table \ref{tab:res_h36m}. For walking and discussion, the model shows better results (3.45 and 3.19 degrees respectively), whereas some tasks like sitting generate a higher mean RMSE of 10.24 degrees. On average, AEPM achieved a mean RMSE of 6.70 degrees, demonstrating robustness across varied movements.

The CMU mocap database further corroborates the effectiveness of AEPM, as shown in Table  \ref{tab:res_cmu}. AEPM yields a mean RMSE of 7.40 degrees for walking and running sequences, while more arduous tasks like navigating uneven terrain result in a mean RMSE of 14.71 degrees. Overall, the average RMSE of 8.73 degrees underlines the generalizability of AEPM to a multitude of real-world scenarios.

\subsection{Time-domain Analysis}

\begin{figure*}
    \centering
    \includegraphics[width=1\linewidth]{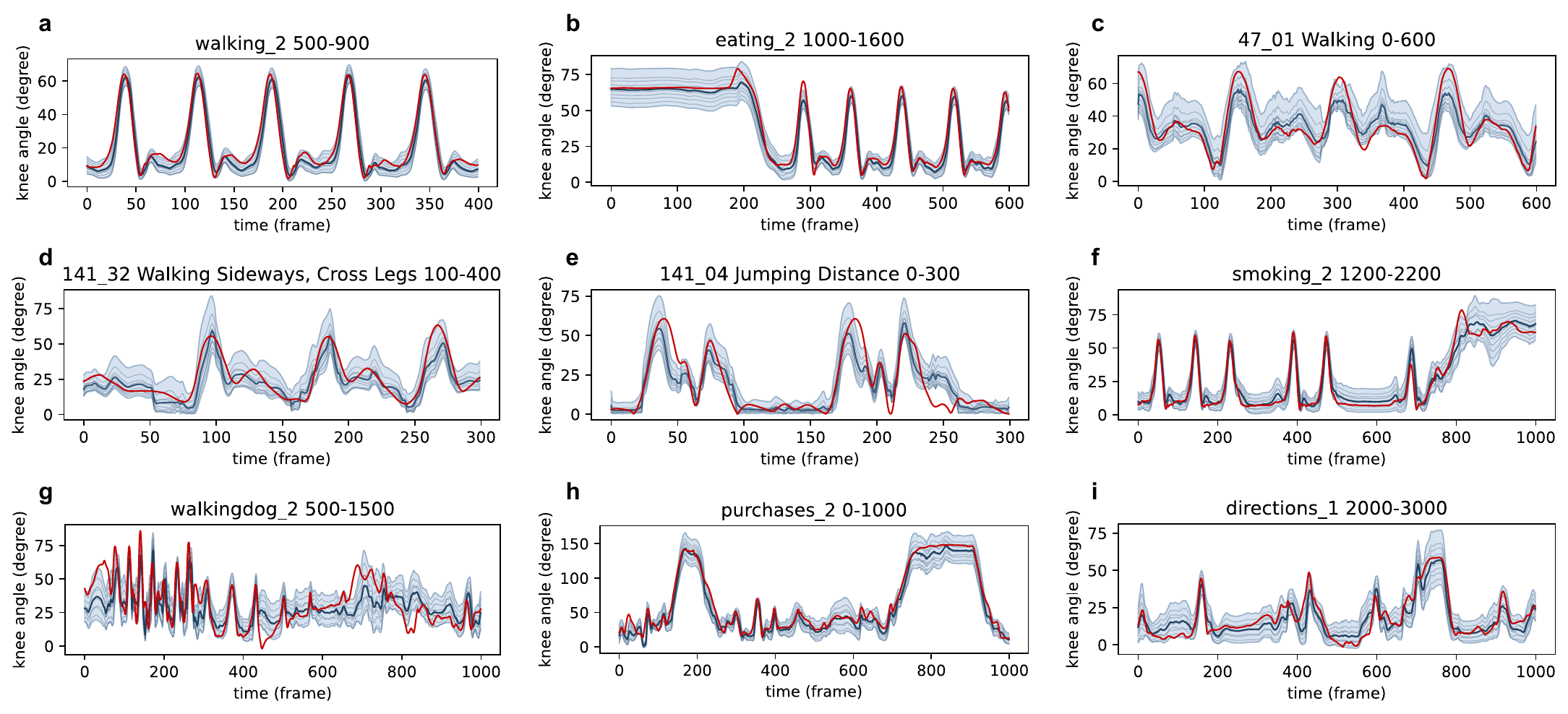}
    \caption{Time-domain results of the knee angle prediction by AEPM. The red line represents the ground truth, the dark blue line represents the mean of predictions and the light blue lines are all the predictions. 
    The graph titles specify the motion type, file reference, and original frame range from the datasets: Human 3.6M (a,b,f-i) and CMU mocap database (c-e). The accuracy of AEPM in estimating knee angles is demonstrated across a spectrum of motion types, including both rhythmic and non-rhythmic activities, as well as walking and other types of motion.
    }
    \label{fig:res_graph}
\end{figure*}

In this section, we will discuss the prediction results of AEPM by three categories: stable rhythmic movement, unstable rhythmic movement and arrhythmic movement. The results are shown in Figure \ref{fig:res_graph}.

\subsubsection{Stable Rhythmic Movement}
Several walking scenarios are shown in the figure, such as Figure \ref{fig:res_graph} (a), (b) and (c). It is shown that AEPM can capture the correct frequency in gait, and estimate the amplitude correctly. Variance during walking is relatively small compared to other scenarios in Human3.6M. While in CMU mocap database, the variance is larger, this might be due to the large variety of gait styles in this dataset, but the estimation still captures the motion and can adjust amplitude correctly even if there are large differences between gait cycles.

Some other rhythmic movements are shown in CMU mocap database, such as \textit{Walking Sideways, Cross Legs} (Figure \ref{fig:res_graph} (d)) and \textit{Jumping Distance} (Figure \ref{fig:res_graph} (e)). In these two cases, the prediction precision is lower, but the model can still correctly capture the frequency correctly.

\subsubsection{Unstable Rhythmic Movement}
We also show some unstable rhythmic movement cases, as is shown in the first half of \textit{smoking} (Figure \ref{fig:res_graph} (f)) and \textit{walking dog} (Figure \ref{fig:res_graph} (g)). In smoking, the subject walked hesitantly, and the model can still precisely capture the movements. In the case of walking dog, the subject is suddenly dragged by the dog. The model fails at the beginning, as the subject is pulled passively and when the subject starts to run, the model still can capture the movements.

\subsubsection{Arrhythmic Movement}
We also examined arrhythmic movement such as \textit{purchases} (Figure \ref{fig:res_graph} (h)) and \textit{directions} (Figure \ref{fig:res_graph} (i)). In the purchase scenario, there is a combination of squatting and standing up, short walking, and upper body movements. In directions, there are short walking, half squatting, and upper body movements. The model can also predict knee movements in these scenarios accurately.

\subsubsection{Transition between Locomotion Modes}
As is shown in Figure \ref{fig:res_graph} (b) and (f), the model can achieve a seamless transition between locomotion modes, or when the situation is even hard to distinguish the locomotion mode, our model can still capture the movement correctly.

Fundamentally, the 0.5-second window is insufficient to encapsulate a complete gait cycle or any other rhythmic movement. Within this brief time window, the model primarily concentrate more on the inter-joint relationships, which will be explored next.

\begin{figure}
    \centering
    \includegraphics[width=1\linewidth]{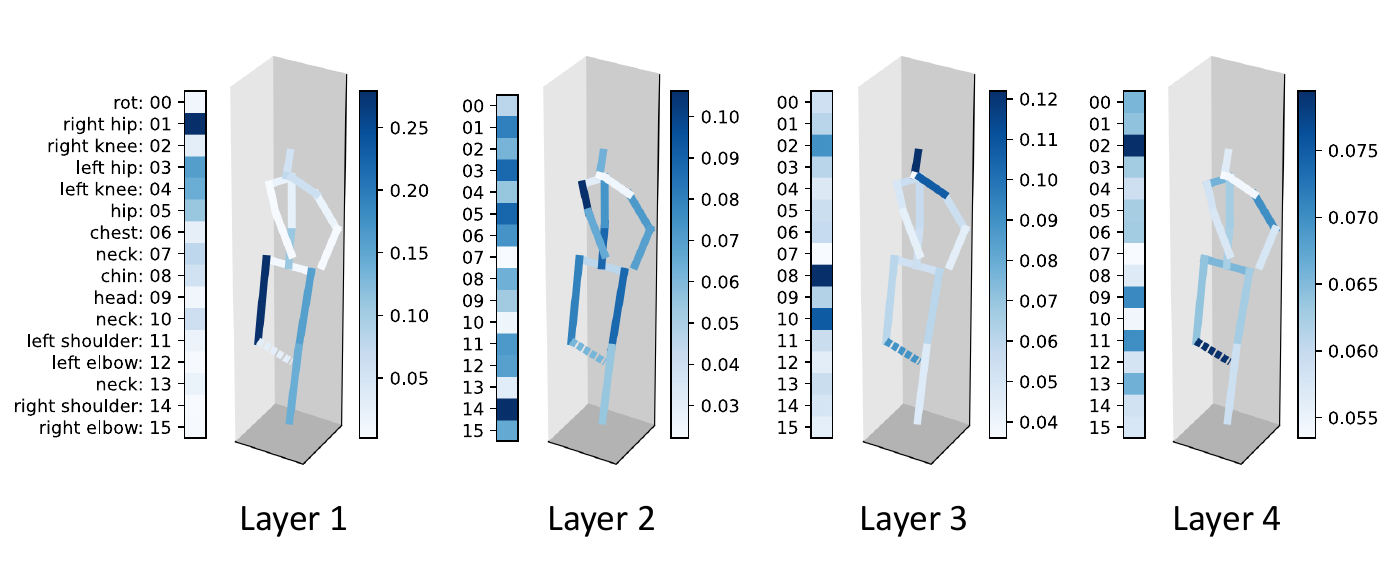}
    \caption{
    The distribution of attention weights across four layers of spatial transformer blocks in a walking scenario in Human 3.6M, with the right knee (masked) designated as the query point. With the progression in deeper layers, there is a discernible shift in attention from joints in close proximity such as the thigh, other leg, and hip, to more distant joints, for instance, the shoulders. Notably, in the final layer, the focus intensifies on the joint under prediction.
    }
    \label{fig:attn_skeleton}
    
\end{figure}

\begin{figure}
    \centering
    \includegraphics[width=1\linewidth]{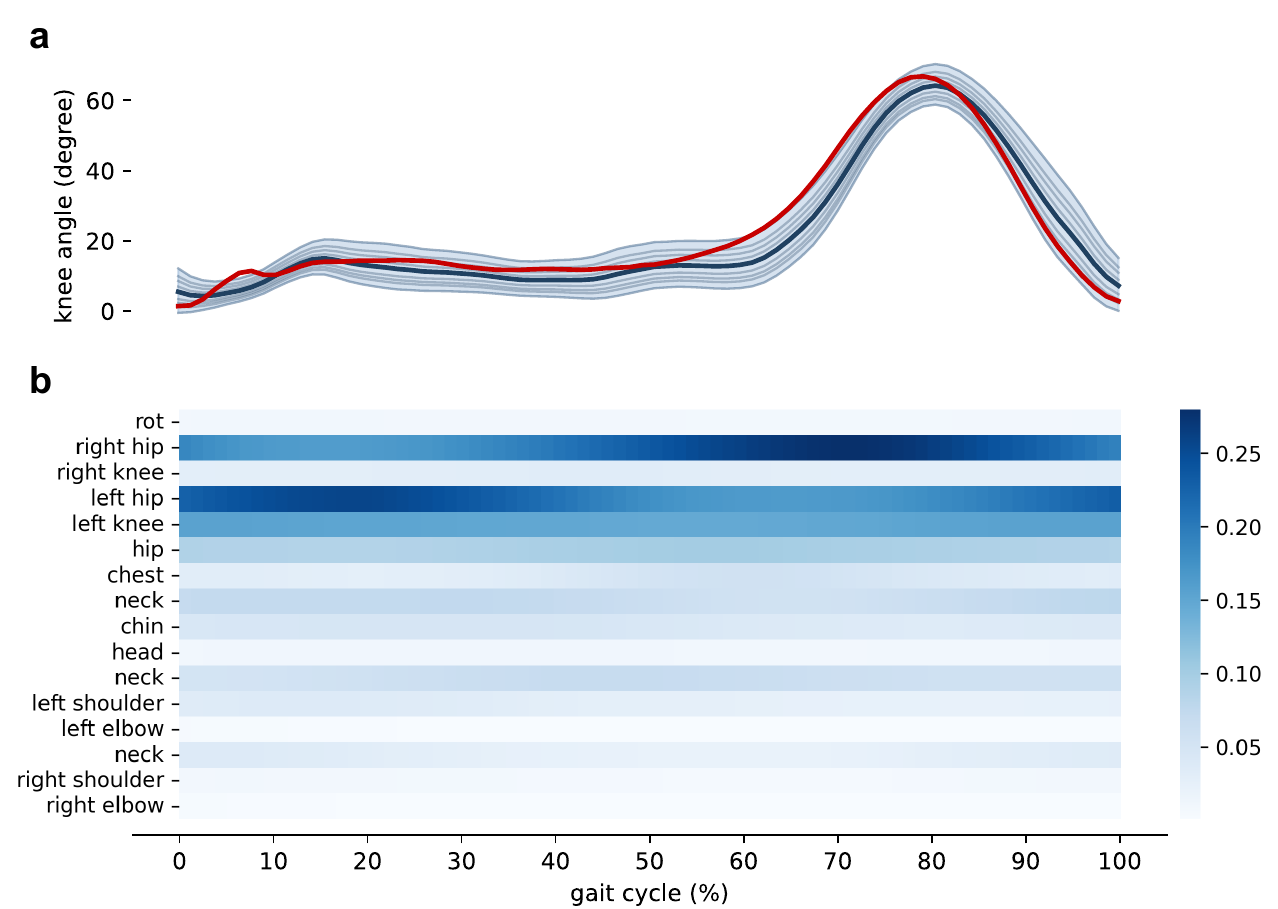}
    \caption{
The distribution and progression of attention weights in the first layers of spatial transformer block in a gait cycle with the right knee (masked) designated as the query point. Graph (a) illustrates the reference of ground truth and predicted knee angle in this gait cycle. Graph (b) shows the attention weights distribution. The relative attention weights remain similar across the whole gait cycle, showing that the basic focus of the model remains similar, while there are also cyclical changes, indicating the model is also adjusting the focus based on different poses in human movement.
}
\label{fig:time_var}
\end{figure}

\subsection{Analyze Joint-wise Synergy}
\label{sec:joint wise}

Human movement and behaviors exhibit high complexity and non-linearity, making them challenging to model statistically. This complexity underpins the choice of the transformer model in this study. The potential relation between joints can also be revealed by checking how the transformer behaves during its prediction.

Transformers use attention mechanisms to weigh the importance of different parts of the input data. The attention weights are computed by queries \(Q\) and keys \(K\) that influence how much emphasis the model places on different parts of the input sequence to produce an output. It allows for adaptive focus, improving the model's ability to extract and leverage pertinent information. Formally, the formula of attention \cite{vaswani2017attention} is given by:
\begin{align}
    \text{Attention}(Q,K,V) = \text{softmax} \left({\frac{QK^T}{\sqrt{d_k}}} \right) V
\end{align}
where \(Q, K, V\) stands for query, key, and value respectively, \(d_k\) is the dimension of keys, and the softmax function is applied to ensure that the weights sum to 1. \(QK^T\) is the attention weights for all queries and keys. In the spatial transformer blocks, the \(Q, K, V\) matrices are all generated by and correspond to joints.

In our analysis, we only visualize the weights whose query is the right knee, the masked one, by taking the corresponding vector \(\mathbf{q}_{r\_knee}K^T\) from \(QK^T\). The attention weights from different layers of STB are visualized, and the weights are averaged through all heads in the same layer and also mapped into the skeleton to show how the model assigns attention with respect to the right knee, as is shown in Figure \ref{fig:attn_skeleton}.

In the initial layer, the model concentrates on adjacent joints affecting knee movement, such as the thigh, contralateral leg, and hip. As the model advances to the second and third layers, attention expands to include the shoulders and back. In the final layer, the focus narrows to the knee itself, indicating that the model has formed an estimation. This progression showcases the model can leverage both simple, direct interactions and complex, global synergy, in a hierarchical manner.

Furthermore, how the attention weights in the first layer evolve throughout a gait cycle is also explored, as is shown in Figure \ref{fig:time_var}. The attention weight remains largely consistent, whereas the importance of the right hip and left hip shifts with the progression of the gait cycle. This suggests a relatively steady interrelation between joints, albeit with some internal variations.

Above all, the spatial attention within AEPM indirectly proves that not only the local anatomical connections, but the entire body's coordination, contains rich information for knee movement.

\subsection{Ablation Experiment Results}
\label{sec:ablation results}
The results of the ablation study are shown in Table \ref{tab:ablation}, where models trained on different losses are compared. As is presented in the table, using \(\mathcal{L}_{S1}\) only has the worst results, and it failed to predict the variance correctly. Using \(\mathcal{L}_{S2}\) shows comparable results with the mixed method, but the mean estimation is a little worse than the two-stage training. Also, the overfitting phenomenon is shown in the early stage of training, leading to a failure to optimize the estimation performance continuously. Applying loss on the knee only has better prediction at best, but much worse in mean, as it tends to predict a much larger variance than loss on all joints.
\begin{table}
    \centering
    \caption{Ablation Experiment Results}
    \begin{tabular}{ccccc}
        \toprule
         &  AEPM&  Loss on knee &  \(\mathcal{L}_{S1}\) only&  \(\mathcal{L}_{S2}\) only \\
         \midrule
         Mean RMSE (degree)&  6.70&  7.55&  7.58&   7.19\\
         Best RMSE (degree)&  2.69&  2.22&  7.56&   2.53\\
         \bottomrule
    \end{tabular}
    \label{tab:ablation}
\end{table}

\subsection{Joint Quantity Experiment Results}
\label{sec:joint quanti}
\begin{table}
    \centering
    \caption{Results of AEPM with Different Input Joint Quantities in Human3.6M Dataset}
    \label{tab:reduce_joint}
    \begin{tabular}{ccccc}
        \toprule
         Joint Quantity&  3&  4&  8& 15\\
         \midrule
         mean RMSE (degree)&  20.28&  11.26&  6.83& 6.70\\
         best RMSE (degree)&  12.75&  4.83&  2.56& 2.69\\
         average predicted std (degree)& 11.84& 8.43& 7.50& 7.12\\
         \bottomrule
    \end{tabular}

\end{table}

The outcome of the joint quantity analysis is shown in Table \ref{tab:reduce_joint}. We assess model performance not only through mean and best RMSE but also by calculating the standard deviation of ten predictions across the entire dataset, serving as an indicator of the model's overall uncertainty. The result delineate that the 8-joint configuration nearly matches the performance of the full 15-joint setting, while a reduction to 4 or 3 joints leads to a marked decrease in accuracy and an increase in standard deviation, highlighting the loss of critical information when excluding upper body joints. This finding suggests that the 8-joint setup strikes a balance between input complexity and estimation accuracy, and indicates that movements and positions of the upper body contribute meaningful information for predicting knee dynamics.

\subsection{Model Comparison}
\begin{table}
    \centering
    \caption{Comparison in Walking Scenario with state-of-the-art Methods}
    \begin{threeparttable}
    \begin{tabular}{cccc}
        \toprule
         \multirow{2}{*}{Methods} & & RMSE  & Sensor\\
         & & (degree)  & Quantity\\
         \midrule
         Eslamy et al. & 2020 \cite{eslamy2020estimation} & 4.5 & 1\\
         Liang et al. & 2021 \cite{liang2021synergy} & 3.89 & 2\\
         Guo et al. & 2023 \cite{guo2023transferable} & 5.56*& 7\\
         Ding et al.& 2023 \cite{ding2023deep} & 10.38 & 7\\
 
         \textbf{AEPM (ours)}& & \textbf{3.45} & 15\\
         \bottomrule
    \end{tabular}
    \begin{tablenotes}    
        \footnotesize              
        \item[*] The RMSE is converted from radius to degree.
    \end{tablenotes}            
    \end{threeparttable}
    \label{tab:comparison}
\end{table}

The AEPM results are compared with other knee angle estimation methods, as shown in Table \ref{tab:comparison}. Our approach is evaluated using walking sequences from the test set of the Human3.6M dataset, with consideration that the CMU mocap database includes a significant volume of unstructured data. The comparative results for other methodologies are directly extracted from existing papers, focusing on level walking scenarios. Despite being trained on a broader dataset that is not solely focused on walking scenarios, our approach demonstrates comparable or even better performance, including the fine-tuned model by Ding et al.\cite{ding2023deep}.

\begin{figure}[h]
    \centering
    \includegraphics[width=1\linewidth]{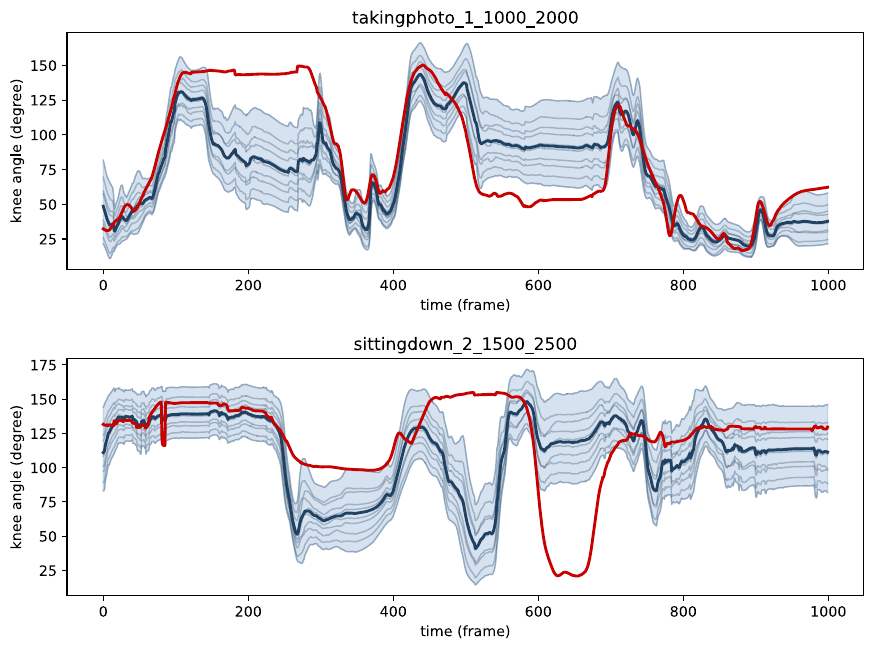}
    \caption{Cases that the model does not show good performance are scenarios such as the person already sitting down and moving legs freely, or the subject is posing or taking photos, where the movement can be unusual. While unable to predict the mean correctly, the model outputs a large variance in these cases, showing that the model can understand these scenarios. Determining the angle in these scenarios might require more explicit intention information.}
    \label{fig:res_failure}
\end{figure}

\subsection{Limitation}

The model tends to underperform primarily in cases where the subject is moved by external forces, or the relation between joints is extremely weak. As for the former one, it can be illustrated in the walking dog scenario (Figure \ref{fig:res_graph} (g)), where the subject is pulled, leading to failed estimations. The latter one is shown in Figure \ref{fig:res_failure}, including two scenarios: sitting and taking photos. In these two cases, the subject moves freely, and the joint angle becomes unpredictable, due to a lack of synergy among joints. However, the output variance also increases in such cases, indicating uncertainty. To address this, incorporating more explicit intention information, such as EMG signals, might be necessary.

Moreover, since the proposed method only focuses on the internal synergy among joints, it cannot access external information. This limitation becomes evident in scenarios such as walking on uneven terrain, as shown in Table \ref{tab:res_cmu}, where the RMSE is twice that of walking and running predictions. However, given that the AEPM model predicts variance, it has the potential to enhance its accuracy by integrating with other methods through information fusion techniques.

\section{Conclusion and Future Work}
In this paper, we propose a novel knee angle estimation method AEPM based on whole-body joint movement. By estimating the mean and variance of the missing knee angle, our method can achieve continuous prediction in action modes beyond the gait cycle and seamless transition between actions. Also, with visualization on attention weights in spatial transformer block, we show that the whole body movements all contribute to the knee prediction.

While our findings provide insights into seamless knee angle estimation, further research is needed to develop an end-to-end version that directly processes sensor data and outputs knee angle to streamline its application. Further, integrating AEPM with additional estimation techniques could address the limitations in discerning user intentions and environmental contexts. Enhancements aimed at predictive capabilities could also be explored, allowing AEPM to forecast movements, and adapt in real-time to dynamic scenarios. Moreover, the study necessitates the recruitment of lower-limb amputees to empirically evaluate the method's efficacy within a real-world application context.

\bibliographystyle{IEEEtran}
\bibliography{reference}

\vfill

\end{document}